\documentclass[10pt,twocolumn,letterpaper]{article}

\usepackage{times}
\usepackage{epsfig}
\usepackage{graphicx}
\usepackage{amsmath}
\usepackage{amssymb}
\usepackage{svg}
\usepackage{subfig}
\usepackage{cite}
\usepackage{url}
\usepackage{titlesec}

 \usepackage{makecell,interfaces-makecell}
\usepackage[breaklinks=true,bookmarks=false]{hyperref}

\begin{document}
\setlength{\parskip}{0pt}
\raggedbottom

\pagenumbering{arabic}

%%%%%%%%% TITLE
\title{DeepDefacer: Automatic Removal of Facial Features via U-Net Image Segmentation}

\date{}

%\author{Anish K. Khazane\\
%{\tt\small akkhazan@berkeley.edu}
%\and
%Julien Hoachuck\\
%{\tt\small jhoachuck3@gatech.edu}
%\and\\
%Krzysztof J.  Gorgolewski
%Stanford University\\
% {\tt\small chris.gor@stanford.edu} 
%\and
%Russell A. Poldrack\\ 
%Stanford University
%Department of Psychology \\  Stanford Center for Reproducible %Neuroscience \\
%Stanford, CA\\
% {\tt\small russpold@stanford.edu}
%}
\author{
\begin{tabular}[t]{c@{\extracolsep{8em}}c} 
Anish Khazane  & Julien Hoachuck \\
{\tt\small akkhazan@berkeley.edu} & {\tt\small jhoachuck3@gatech.edu}\\ 
\\
\\
Krzysztof J.  Gorgolewski & Russell A. Poldrack\\
Department of Psychology &  Department of Psychology\\
Stanford University & Stanford University
\end{tabular}
}
\maketitle
%\thispagestyle{empty}

%%%%%%%%% ABSTRACT
\begin{abstract}
   Recent advancements in the field of magnetic resonance imaging (MRI) have enabled large-scale collaboration among clinicians and researchers for neuroimaging tasks. However, researchers are often forced to use outdated and slow software to anonymize MRI images for publication. These programs specifically perform expensive mathematical operations over 3D images that rapidly slows down anonymization speed as an image's volume increases in size.

   In this paper, we introduce DeepDefacer, an application of deep learning to MRI anonymization that uses a streamlined 3D U-Net network to mask facial regions in MRI images with a significant increase in speed over traditional de-identification software. We train DeepDefacer on MRI images from the Brain Development Organization (IXI) and International Consortium for Brain Mapping (ICBM) and quantitatively evaluate our model against a baseline 3D U-Net model with regards to Dice, recall, and precision scores. We also evaluate DeepDefacer against Pydeface, a traditional defacing application, with regards to speed on a range of CPU and GPU devices and qualitatively evaluate our model's defaced output versus the ground truth images produced by Pydeface. We provide a link to a PyPi program at the end of this manuscript to encourage further research into the application of deep learning to MRI anonymization.
   
\end{abstract}

%%%%%%%%% BODY TEXT
\section{Introduction}

Within the past few decades, improvements in computing technology have allowed researchers and clinicians to create more detailed magnetic resonance imaging (MRI) images 
and store gigabytes of sensitive information in electronic patient databases \cite{privacydefacing}. However, the growing storage and dissemination of MRI images raises privacy concerns as neuroimaging data can reveal sensitive facial information about patients, as shown in Figure 1. \cite{HIPAA}.
The U.S Health Insurance Portability and Accountability Act (HIPAA) defines any full face photographic images as protected health information (PHI), and mandates the de-identification of sensitive information prior to sharing of the data \cite{HIPAA}. Thus, the goal of de-identification is to remove any identifying facial information in a MRI image.

\begin{figure}%
    \centering
    \vspace{0pt}
    \subfloat[]{{\includegraphics[width=7.7cm]{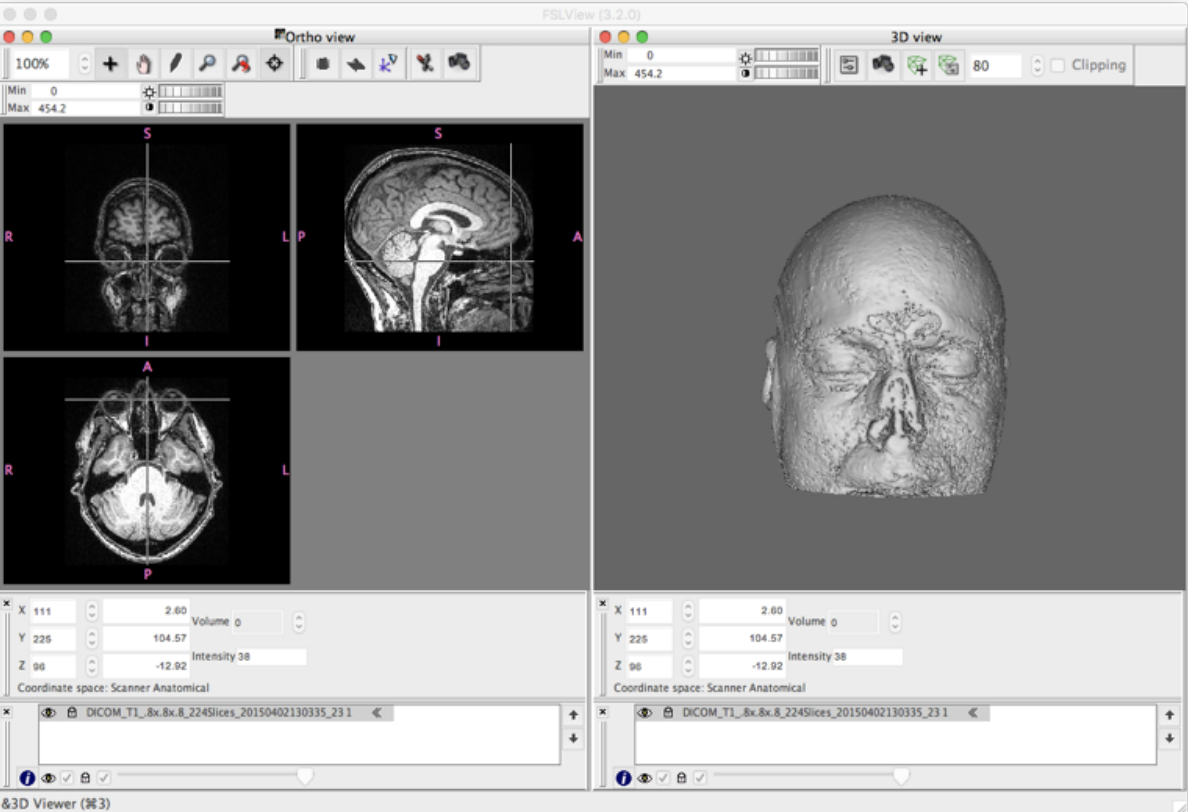} }}%
    \caption{An example of a 3D reconstructed image from a MRI scan \cite{poldrack}.}%
    \label{fig:example}%
    \setlength\belowcaptionskip{-1ex}
\end{figure}

Several standard methods attempt to tackle de-identification by using linear co-registration methods to mask non-brain tissue in an image \cite{defacing}. "Pydeface", a common de-identification software, defaces an image by using co-registration methods to translate a facemask onto the facial region of a scan.  \cite{defacing}. Other techniques use a "facial atlas," to predict the probability that a voxel represents facial features \cite{privacydefacing}. However, these techniques are limited because the constructed facial atlas can only be applied to T1-Weighted MRI images, which encompasses a very limited range of voxel intensities. Pydeface, on the other hand, does support co-registration using between-modality cost functions - e.g mutual information (MMI), normalized MMI - to decrease the error between intensity values in a reference image and an input image of a different contrast type \cite{pydeface}. However, images with small resolution sizes can throw these correlation methods off balance as many cost functions rely on constructing 2D joint intensity histograms that are especially susceptible to differences among a small number of voxels.

In this paper, we present Deepdefacer, a modified 3D U-Net image segmentation network that can rapidly deface 3D MRI scans on both CPU and GPU devices. To our knowledge, our work is the first application of deep learning to MRI anonymization.

Due to having a fully convolutional architecture that is also heavily streamlined, Deepdefacer has 92\% fewer parameters than the original 3D U-Net model and is able to deface MRI images at nearly 90\% faster speed than Pydeface. We train the network on pairs of non-defaced MRI images and 3D binary masks from the Brain Development Organization (IXI) and the International Consortium for Brain Mapping (ICBM) datasets in order to predict binary masks that are subsequently multiplied against original MRI images to deface them. \cite{IXI, ICBM}. We use a traditional defacing software, Pydeface, to construct the ground truth masks in our input. Figure 2 shows examples of images in our datasets.

\begin{figure}%
    \centering
    \vspace{0pt}
    \subfloat[]{{\includegraphics[width=8cm]{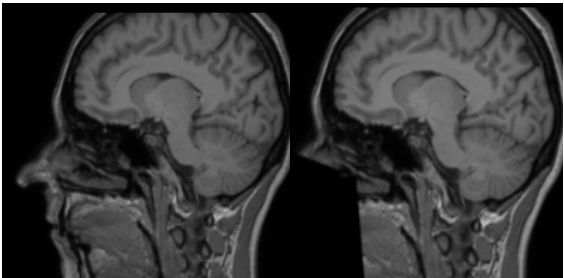} }}%
    \caption{Example of a T1-Weighted MRI scan on the left and a proper defacing of the image on the right.}%
    \label{fig:example}%
    \setlength\belowcaptionskip{-1ex}
\end{figure}

In the following sections, we briefly expand on the disadvantages of classical de-identification software (such as Pydeface) that use numerical algorithms to anonymize MRI images. We then discuss 2D versus 3D image segmentation networks for medical imaging and our reasoning behind using the 3D U-Net architecture for this task. Afterwards, we present our own modifications to the 3D U-Net network to create Deepdefacer, along with explaining its full training and inference pipeline. We then quantitatively compare the model against a baseline 3D U-Net architecture with regards to Dice, precision, and recall scores and Pydeface for speed. We also qualitatively compare outputs from both Deepdefacer and Pydeface to analyze model generalization. Lastly, we conclude this paper by discussing areas where Deepdefacer's performance can be further improved along with plans for building a product API that allows users of all technical backgrounds to utilize the model for MRI defacing.

\section{Related Works}
\subsection{Classical De-Identification Algorithms}
Classical algorithms for MRI anonymization typically fall into two categories: skull-stripping and co-registration \cite{skull}. Skull-stripping techniques delineate the brain boundary of a MRI scan in order to completely remove voxels representing non-brain tissue from the original image. While these stripping methods can adequately remove facial features in scans, their performance is typically conditional on the MRI image type (T1-weighted, T2-weighted, etc) and consequently may incorrectly remove voxels representing brain tissue if image composition is previously unknown to the program \cite{skull}. Furthermore, several skull-stripping techniques are highly sensitive to differences in image resolution and size, are unable to generalize across data sets, and can incorrectly modify images in ways that impair post processing and analysis. \cite{privacydefacing,defacing}.  

Co-registration techniques fall into two different categories: automated and manual registration. Pydeface, an automated registration tool, calculates an affine transformation from a reference template to an input image, which is subsequently applied against a facemask to translate the map over the facial region of the original input image \cite{pydeface}. As previously mentioned, Pydeface also supports between-modality registration which uses appropriate cost functions to determine an optimal linear transformation for images with unique compositions. One of Pydeface's critical shortcomings however is anonymization speed: Pydeface in particular takes upwards of several minutes to deface images with high resolution. On the other hand, manual registration techniques employ user-defined landmarks on the brain portion of a scan to help registration software define an affine transformation that properly defaces an image \cite{coregistration}. While pre-defined landmarks improves defacing speed, the time-consuming - and subjective - task of manually modifying MRI scans with markers is not a scalable option \cite{coregistration}. 

\subsection{Survey of Image Segmentation Networks}
To avoid the shortcomings of classical de-identification algorithms, we turn to deep learning and survey the computer vision literature on image segmentation for medical imaging. Our goal is to construct a neural network that can   correctly and rapidly anonymize images with similar voxel compositions to the scans that the model was trained on. Convolutional neural networks (CNNs) are particularly adept at learning complex spatial and intensity patterns in an image, where some of these patterns might be the curvature and edges that make up facial features in a MRI scan \cite{surveyCNN}. Fully convolutional neural networks (FCNs) and U-Net image segmentation networks are two commonly used networks for semantic segmentation in medical imaging \cite{surveyCNN}. 

FCNs, as implied by their name, have a fully convolutional architecture that allows the network to train and predict on variable sized images \cite{fcn}. However, FCNs send an input image through several alternating convolutional and pooling layers that rapidly downsample the image's resolution prior to prediction in its last layer \cite{fcn}. The final output of a FCN is typically low resolution and consequently may contain inaccurate boundaries between features in a scan. Furthermore, FCNs have primarily only been used for 2D image segmentation because there are very few stable implementations of 3D FCNs for medical image segmentation \cite{3DUNet}.

On the other hand, there are widely adopted 2D and 3D implementations of U-Net image segmentation networks. Similar to a FCN, the U-Net is fully convolutional and has a contracting path at the beginning of the network that captures contextual information within images via compact feature maps \cite{3DUNet}. However, a symmetric upsampling path immediately follows this contraction phase to retain boundary information that might have been obfuscated at the beginning of the network \cite{3DUNet}. This key difference between the U-Net and FCN gives the former network an edge with regards to semantically segmenting medical images with intricate details like MRI images. The fact that all of the U-Net's layers are pooling or convolutional also makes the network very computationally efficient due to having far fewer parameters than other segmentation networks. Consequently, the U-Net's superior speed and ability to precisely localize boundaries in volumetric image data makes it a particularly attractive tool for masking 3D MRI images.

\subsection{2D U-Net}
The 2D U-Net is a commonly used image segmentation network in medical imaging. Consisting of multiple 2D CNNs, this model typically does not suffer from high computational or memory costs, because it operates over planar regions rather than volumetric data \cite{unet}. Furthermore, one can use a 2D U-Net to segment images in 3D space by simultaneously training multiple 2D U-Nets on orthogonal projections of a 3D image. This distributes the required computational power for defacing a 3D image across multiple networks and could rapidly improve anonymization speed over classical defacing software \cite{triplanar}. 

While a parallelized 2D U-net implementation could save on speed, memory, and computational costs, there are caveats with 3D MRI data that make this network an unsuitable choice for our task. Firstly, MRI data in particular is very sparse with many background voxels that obviously are not related to the primary image in the scan. Thus, a 2D U-Net which cannot use spatial information while defacing slices is especially susceptible to sections with little contextual information. We find that these highly unbalanced slices in a 3D MRI scan are especially difficult for a 2D U-Net to mask, and consequently this network performs worse than a 3D U-Net approach with respect to fully anonymizing an image.

\subsection{3D U-Net}
Within the taxonomy of U-Net architectures, a 3D U-Net is essentially identical to a 2D U-Net other than convolving over volumetric data as opposed to flat data \cite{3DUNet}. However, as previously mentioned, an extra dimensional axis provides a 3D U-Net with critical spatial information that helps the network learn more robust boundaries while segmenting images. In fact, 3D image segmentation has quantitatively performed better than 2D segmentation for other medical imaging tasks in relevant literature \cite{2dvs3d}. 

However, it is important to note that the lack of sufficiently large medical imaging datasets for training 3D U-Nets is a significant disadvantage compared to training 2D U-Nets, which can simply transform a single volumetric image into thousands of 2D input slices. One particular 3D U-Net implementation by Cicek et al. tackles this issue by performing on-the-fly data elastic deformations of 3D data for data augmentation during training \cite{2dvs3d}. This addition heavily improves the 3D U-Net's generalizability, and we borrow several key data augmentation insights from the author's implementation for training our own modified 3D U-Net for Deepdefacer. Furthermore, we stick with an underlying 3D U-Net architecture for MRI anonymization because while the lack of sufficient image data is an issue for 3D networks, we change our optimization objective (as detailed in the Proposed Approach section) to learn 3D binary masks rather than directly deface images. Consequently, training Deepdefacer to only predict binary facemasks rather than segment intricate boundaries between facial features in a scan is a far simpler objective and is less dependent on a large training dataset.

\setlength\belowcaptionskip{-3ex}
\begin{figure*}
    \includegraphics[width=\textwidth, height=5cm]
               {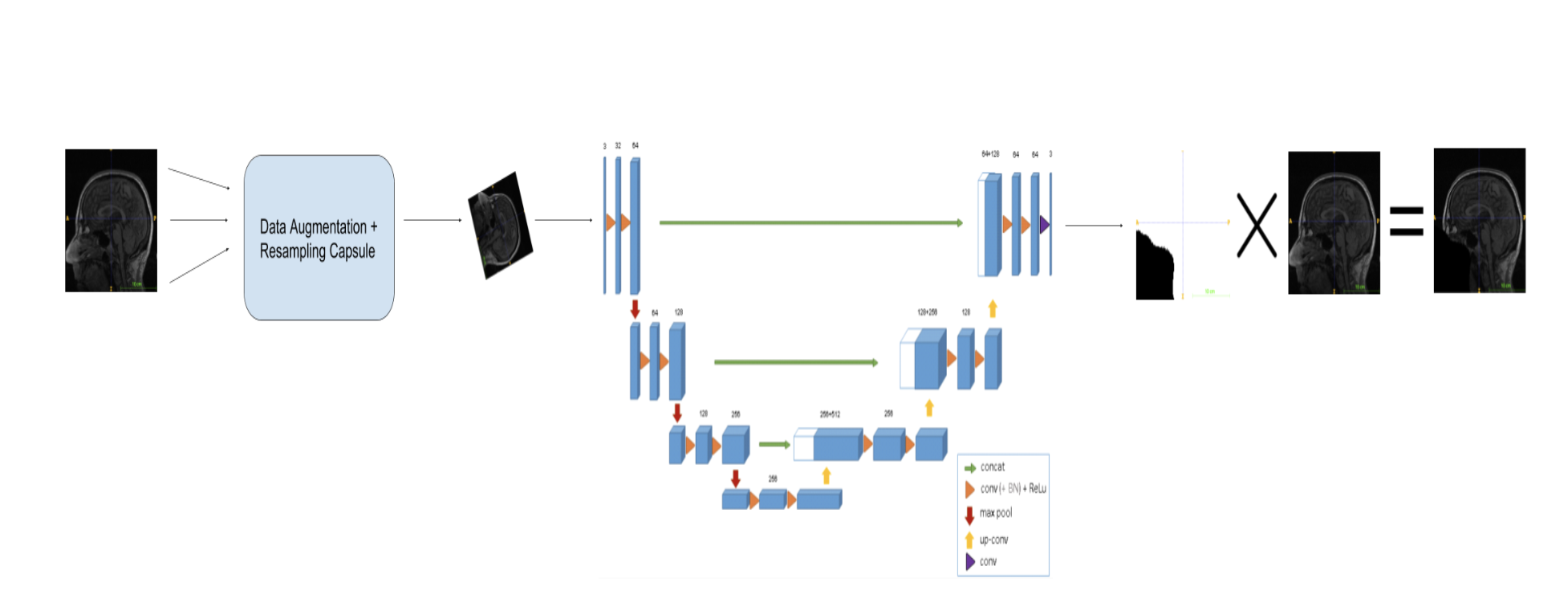}
    \caption{Deepdefacer's training pipeline. Changing the 3D U-Net's optimization objective, adding a pre-processing module, and using fast matrix operations rapidly improves defacing speed.}%
    \label{fig:example}%th
\end{figure*}

\section{Proposed Approach}

Motivated by the success of U-Net image segmentation networks in medical imaging, we propose Deepdefacer, a modified 3D U-Net model that can rapidly anonymize MRI images. We begin this section by first detailing the baseline 3D U-Net's architecture, which is optimized to directly deface input MRI images. Afterwards, we present Deepdefacer's own modified 3D U-Net architecture.

\subsection{3D U-Net Baseline}

The baseline 3D U-Net segmentation network includes a downsampling and upsampling stage, along with a softmax output in its last layer to directly segment the facial region from the brain tissue within a MRI image.

Each of the four convolutional blocks in the downsampling stage contains two 3D convolutional layers with 3 x 3 x 3 sized kernels with ReLU activations. These layers are followed by a 3D max pooling layer that halves all dimensions with a stride of 2. The max pooling layer is followed by a batch normalization layer. All blocks in both the upsampling and downsampling phases use (32, 64, 128, 1024) filter sizes as in the original 3D U-Net architecture \cite{3DUNet}.

The second stage incorporates four upsampling blocks. Each block includes a 3D convolutional layer of kernel size 3 x 3 x 3 followed by a 3D upsampling layer of size 2 in each dimension. The output of the upsampling layer is concatenated symmetrically with the output of the downsampling block that has the same number of filters \cite{3DUNet}.

The 3D U-Net's last layer is a 1 x 1 x 1 convolutional layer that produces a MRI image of the same size as the original scan but with a defaced region. This model uses a softmax cross-entropy loss to measure the error between a predicted defaced MRI image and the ground truth defaced MRI image. 

In order to compare the quality of this model's defaced output to that from Deepdefacer, we subtract the baseline model's predicted defaced MRI from the original MRI image and then threshold the subtracted volume to produce a binary segmentation map made up of 0s and 1s. We then compare this map to the binary masks produced by Deepdefacer for both quantitative and qualitative evaluation.

\subsection{Deepdefacer: 3D U-Net for 3D Binary Mask Prediction} 

Deepdefacer's primary objective is to improve MRI anonymization speed while creating high-quality masks that rival output - on images of similar voxel composition - from more specialized programs like Pydeface. Consequently, we use an underlying 3D U-Net as the base architecture for Deepdefacer, but make the modifications listed in this section to create a more optimized network for our task. Deepdefacer's full training pipeline is depicted in Figure 3.

\subsubsection{Optimization Objective: Binary Cross Entropy Loss} 

We modify the 3D U-Net's optimization objective to instead reduce the binary cross entropy loss between a predicted 3D binary mask and a label map, as shown in equation (\ref{bce}), where $N$ is the number of MRI images in a batch of data, and $y_n$ and $1 - y_n$ represent two class values \{0,1\} for any predicted voxel in the output mask. 

\begin{equation}\label{bce}
BCE = -{\frac {1}{N}}\sum _{n=1}^{N}\ {\bigg [}y_{n}\log {\hat {y}}_{n}+(1-y_{n})\log(1-{\hat {y}}_{n}){\bigg ]}
\end{equation}

Accordingly, we pass the output of the last convolutional layer in a 3D U-Net through a sigmoid activation function to predict 3D binary masks. A 0 in a 3D binary mask represents a region to deface whereas 1 represents a region to maintain. $\hat{y_n}$ and $1 - \hat{y_n}$ represent the model's probability of predicting a 1 or 0, respectively, for a particular voxel in the output mask, of which we compute a binary cross-entropy loss by using the logarithmic functions depicted in (\ref{bce}).

\subsubsection{Model Architecture: Reduced Filter Sizes and Layers} 

We take advantage of our simpler optimization objective by heavily reducing the filter sizes in each layer of the original 3D U-Net by a factor of 4  (8, 16, 32, 64, 128) to boost computational efficiency and prevent Deepdefacer from producing overfitted binary masks. We also remove the batch normalization layers from the original 3D U-Net architecture to avoid worsening issues related to high class imbalance in MRI images: the high number of black voxels (0 intensity) in MRI images typically dwarfs the percentage of voxel intensities pertaining to the mask in the dataset. We find that these intensities negatively impact the global mean and deviation calculated in batch normalization layers and consequently cloud the quality of this model's mask output. Making the two aforementioned changes results in a 92\% reduction in the total number of model parameters in comparison to the original 3D U-Net. As we'll explore in greater detail in the Analysis section of this paper, we partly attribute our model's fast inference speed to this massive reduction in model size.

\subsubsection{Pre-Processing Stages: Data Augmentation and Resampling} 

\begin{table*}[t]
\begin{center}
 \begin{tabular}{||c | c |  c | c||} 
 \hline
  Source & Voxel Size [mm] & Field of View & Number of Training Samples \\ [0.5ex] 
 \hline\hline
 IXI & 0.62, 1.0, 1.0 & 256x256x150 & 140 \\
 \hline
 IXI & 0.66, 1.0, 1.0 & 256x256x150 & 43 \\ 
 \hline
 IXI & 0.8, 0.8, 0.8 & 256x256x146 & 146 \\ 
 \hline
 ICBM & 1.24, 0.98, 0.98 & 256x320x256 & 242\\ 
 \hline
 ICBM & 1.25, 1.25, 1.20 & 174x256x256 & 101\\ 
 \hline
 ICBM & 1.3, 0.98, 0.98 & 220x320x208 & 54 \\ 
 \hline
\end{tabular}
   \captionof{table}{Sample of images from the training set that vary in voxel size, field of view, and image source. The entire training dataset includes 20 unique protocols. }\label{table:dataset}
   \vspace{10pt}
\end{center}
\end{table*}

We also add key pre-processing stages to Deepdefacer's training pipeline to improve the quality of its predicted binary masks. Inspired by elastic deformation techniques used by Cicek et al. \cite{2dvs3d} for regularizing 3D segmentation networks, we add a data augmentation capsule to our training pipeline that spatially rotates and scales training images prior to running them through the network. Performing this data augmentation on-the-fly turns a training set of roughly 2,000 3D MRI images into over 10,000 with unique volumetric orientations, which bolsters Deepdefacer's ability to predict masks that generalize to many different types of MRI images. We also add a resampling layer that reduces an input image's dimensions to a much smaller size prior to running it through the network, and then resamples the map back up to the original image size after prediction. This portion of our pipeline allows us to heavily constrain the space complexity that comes with predicting masks against massive MRI images. This layer is especially critical for creating a tool that rapidly anonymize input on commercial CPU devices. 

\subsubsection{Post-processing Stages: Hadamard Product and Thresholding} 

After the mask has been resampled back to the original image size, we threshold it to discrete values [0,1] to create a clean binary map. We specifically use a threshold value of 0.5, and determined this quantity with the following protocol: run a log linear search between [0,1] to retrieve threshold values, process output masks with each of these quantities, then test these maps against a held-out validation set to determine an optimal value for thresholding.

Post thresholding, we multiply the mask against the original image data to fully deface it. We use the Hadamard product in equation (\ref{Hadamard Product}) for this multiplication, where $A$ is a binary matrix and $B$ is voxel data representing the original MRI image:

\begin{equation}\label{Hadamard Product}
(A {\circ} B)_{ij} = (A)_{ij}(B)_{ij}
\end{equation}

We believe that changing the 3D U-Net's objective to predicting a binary map allows Deepdefacer to more easily generalize to different types of MRI images instead of overfitting on intricate anatomical features within any particular scan. Furthermore, multiplying the mask against the original image as a post-processing step allows us to maintain the quality of all sections within the image that are not in the facial region. We will see in the Experiments section that directly segmenting a MRI image can incorrectly perturb voxel data that represents critical information like brain tissue.

\begin{figure}[hbp]
\begin{minipage}[t]{0.45\linewidth}
    \includegraphics[width=\linewidth]{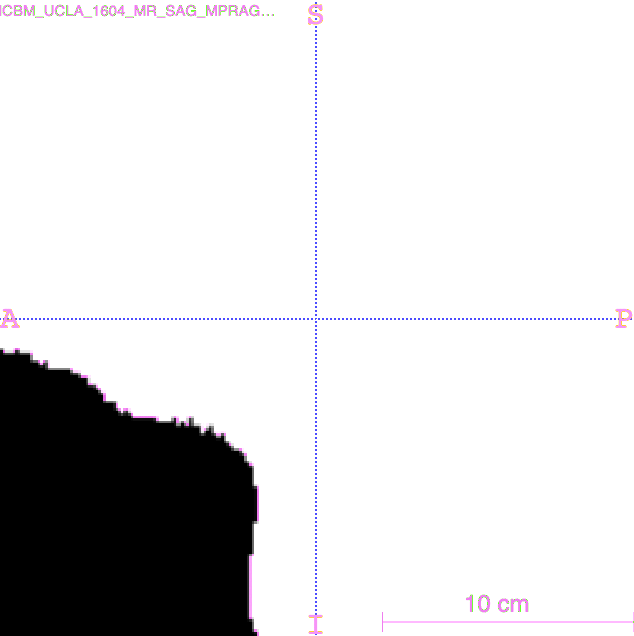}
    \caption{Output mask from Deepdefacer. A (Anterior), S (Superior), P (Posterior) and I (Inferior) are directional axes defining the image's orientation.}
    \label{f1}
\end{minipage}%
    \hfill%
\begin{minipage}[t]{0.45\linewidth}
    \includegraphics[width=\linewidth]{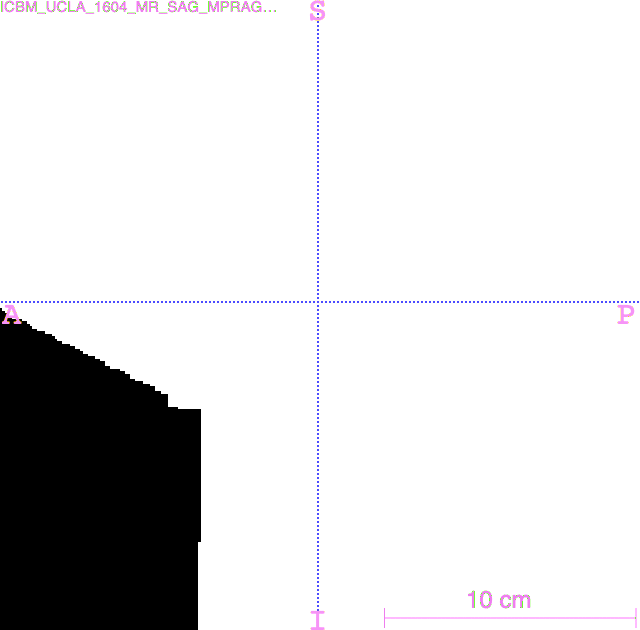}
    \caption{Ground truth mask from Pydeface. Axes follow the same definition listed under Figure 4.}
    \label{f2}
\end{minipage} 
\end{figure}

% \begin{table*}[t]
% \begin{center}
%  \begin{tabular}{||c | c |  c | c ||} 
%  \hline
%   & Intel i7 CPU (s) & 4 vCPUs 15 GB (s) & 1 Nvidia Tesla K80 (s) \\ [0.5ex] 
%  \hline\hline
%  Deepdefacer & \textbf{20.707 (-90\%)} & \textbf{49.779 (-68\%)} & \textbf{13.527 (-91\%)} \\ 
%  \hline
%  3D U-Net Baseline & NA (Crashed) & 85.026 & 30.524 \\
%  \hline
%   Pydeface & $225.249$ & 157.089 & 157.089 (No GPU benefit) \\ [1ex] 
%  \hline
% \end{tabular}
%     \captionof{table}{Average speed test results across a commercial Intel i7 2.8GHz chip, 4 VCPUs, and 1 Nvidia Tesla K80 GPU.}\label{table:speed}
% \end{center}
% \end{table*}

\section{Implementation Details}

\subsubsection{Data}

Our data consists of 1928 T1-weighted MRI images from the Brain Development Organization (IXI) and International Consortium for Brain Mapping (ICBM). As Deepdefacer has four downsampling and upsampling stages, all T1-weighted images must have dimensions that are divisible by 16. Consequently, we resample the  $HxWxDxC$ resolution sizes of IXI MRI images to 256x256x150x1, with a single channel $C$ to capture black and white voxel values. We resample images from ICBM to match a 256x320x256x1 resolution size. We use linear interpolation for both resampling tasks.

We also create a label mask for each image by passing the input MRI images through Pydeface to construct segmentation maps that will be used as ground truth during training. Figure. 4 and 5 shows examples of a predicted mask from Deepdefacer, and a ground truth mask from Pydeface respectively.

Furthermore, we split our data into training, validation, and test sets based on the underlying voxel composition for all MRI images in our dataset. We have 30 unique voxel sizes in $mm$ space across both IXI and ICBM, and we use a 80\%, 10\%, and 10\% split for experimentation; this amounts to 1542 images and 20 voxel spacing protocols in the training set, 192 images and 5 voxel spacing protocols in the validation set, and 192 images and 5 voxel spacing protocols in the test set. Table \ref{table:dataset} provides a sample of different configurations in the training set. Additionally, we normalize input MRI images to a [0,1] range to match Deepdefacer's sigmoid output distribution to promote training convergence.

\subsubsection{Training}

We train Deepdefacer with 2 Nvidia K80 GPUs for 4000 training iterations (3 epochs) on 1542 MRI images, which takes roughly 3 days. We specifically train with Adam optimization with a learning rate, $\beta_1$, and $\beta_2$ values of $1e^{-4}$, 0.9, and 0.999 respectively.

We use a batch size of 1 to maintain computational efficiency while training Deepdefacer on large 3D MRI images. However, we speed up training by using memory efficient cudaNN convolution layers within our network. 

Furthermore, we initialize all weights with He normalization to stabilize the gradient backpropagation of ReLU units in each convolutional layer.

\section{Experiments and Results}

\subsection{Quantitative Results} 
In order to exhaustively evaluate both the quality and speed of our defacing models, we choose the following quantitative measures for comparison: speed, Dice, precision, and recall. All measures were computed over predictions from a test set of MRI images with different voxel spacing protocols than the training set. We compute Dice, precision, and recall scores for Deepdefacer by specifically comparing its predicted 3D binary masks with corresponding respective ground truth 3D masks from Pydeface. As previously mentioned, we compare the 3D U-Net baseline with Deepdefacer by subtracting the baseline's segmented MRI outputs from the original MRI scans. We then threshold the delta map to only discrete values [0,1] to yield a binary segmentation map for quantitative comparison.

\begin{figure}%
    \centering
    \vspace{0pt}
    \subfloat[]{{\includegraphics[scale=0.55]{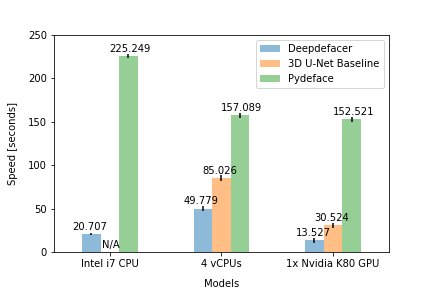} }}%
    \caption{Average speed test results across commercial CPU and GPU devices. The 3D U-Net baseline model crashes while running on an Intel i7 CPU, and Pydeface does not significantly benefit from running computations on GPU.}%
    \label{fig:speed}%
    \setlength\belowcaptionskip{-1ex}
\end{figure}

\begin{table*}[t]
\begin{center}
 \begin{tabular}{||c | c |  c |  c| c||} 
 \hline
  & Dice & Precision & Recall & Number of Model Parameters \\ [0.5ex] 
 \hline\hline
 Deepdefacer & \textbf{0.854} & \textbf{0.916} & 0.805 & \textbf{1,412,197 (-92\%)} \\
 \hline
 Base 3D U-Net & $0.413$ & $0.132$ & \textbf{0.882} & 19,069,955 \\ 
 \hline
\end{tabular}
   \captionof{table}{Deepdefacer VS 3D U-Net Baseline with respect to common binary segmentation metrics and model size.}\label{table:prdice}
   \vspace{10pt}
\end{center}
\end{table*}

\subsubsection{Speed}
We test Deepdefacer, the 3D U-Net baseline model, and Pydeface on different CPU and GPU devices to record the average time taken to predict all images in the test set. We assess all three models with an array of different devices: an Intel i7 CPU, 4 vCPUs with 15 GB of memory on a Google Cloud server, and 1 Nvidia Tesla K80 GPU card. Figure \ref{fig:speed} provides the results for all three of these speed tests.

From analyzing Figure \ref{fig:speed}, we see that both U-Net models are able to deface input MRI images significantly faster than Pydeface. However, Deepdefacer is able to deface images 90\% faster than Pydeface on a commercial Intel CPU card whereas the baseline 3D U-Net model requires at least 4 vCPUs with 15GB total memory to store computations from masking a single image. Deepdefacer also outperforms the other two models on vCPUs by upwards of 68\% and a typical server GPU by 91\%.

We attribute Deepdefacer's massive speedup over the 3D U-Net baseline to a variety of significant factors: a far smaller model size, critical resampling module, and simpler task of predicting masks rather than directly segmenting images. Table \ref{table:prdice} also shows that Deepdefacer has roughly 92\% fewer parameters than the 3D U-Net baseline due to a smaller filter size per layer ratio, which not only saves computational resources during training and inference but also allows Deepdefacer be stored on compact commercial devices. Furthermore, resampling images to smaller input resolution sizes saves additional space complexity in comparison to the 3D U-Net baseline's strategy of directly segmenting massive MRI images. Predicting binary maps rather than directly segmenting MRI images also rapidly decreases the computational complexity of the model.

Along with the reasons stated above, we also attribute Deepdefacer's speedup to Pydeface's own inefficiencies while masking images. For example, Pydeface's co-registration pipeline builds an affine transformation matrix based on a template for any image, which can be an an expensive mathematical procedure depending on the total number of voxels in a given input image. For example, performing an affine transformation over 9,830,400 voxels in a 256x256x150 MRI image is computationally expensive and the lack of a resampling module is a major reason why the software is slower in comparison to Deepdefacer. In addition, Deepdefacer's cudaNN optimized layers also allows the network to a fuller advantage of GPU resources, which only further improves its prediction speed in comparison to Pydeface.

\subsubsection{Dice Coefficient} 
    
The most common and if not de-facto measure used for evaluating binary segmentation tasks is the Dice coefficient.
The Dice coefficient measures the size of the overlap between a predicted binary segmentation mask $X$ and the ground truth segmentation output $Y$ as depicted in equation (\ref{dice}). For MRI anonymization, the score penalizes a model for falsely segmenting non-facial region voxels (false positives), and for not segmenting facial region voxels (false negatives). We present the results from this experiment in Table \ref{table:prdice}.

\begin{equation}\label{dice}
Dice = \frac{2|X\cap Y|}{|X|+|Y|}
\end{equation}
\vspace{5pt}

As expected, the 3D baseline model performs far worse than Deepdefacer because the former model attempts to directly segment an entire image end-to-end rather than construct a simple binary mask during training. Figure 8 shows examples of segmentation maps from both Deepdefacer and the 3D U-Net baseline that reinforce this observation. The baseline model's segmentation map in (b) incorrectly perturbs several voxel intensities in non-facial regions of the MRI image, which implies a greater number of false positives and consequently a lower Dice coefficient score. On the other hand, Deepdefacer's map in (a) does not disturb areas outside of the facial region of the scan, and consequently has a larger Dice coefficient score in comparison to the baseline model. This is likely because Deepdefacer's objective function is far simpler to optimize. By training to predict 3D binary masks, Deepdefacer can ignore misleading patterns from anatomical features, low or high voxel intensity values, and other features of a MRI scan and only focus on masking the general location of an image's facial region. 
\subsubsection{Precision and Recall} 
We also compare Deepdefacer with the baseline U-Net model with respect to both precision and recall scores.  Precision allows us to compute a ratio of true positives to false positives, while Recall allows us to compute a ratio of true positives to false negatives. 

As shown in Table. \ref{table:prdice}, the baseline model has a very high recall and low precision score. This implies that while the baseline model classifies a substantial number of voxels within a MRI scan as a part of a mask, it is unable to maintain the quality of the rest of the image. On the other hand, Deepdefacer balances a relatively high recall score of 0.805 with a high precision score of 0.916, implying that this model is able to better constrain the number of defaced voxels to an image's facial region. Consequently, Deepdefacer is able to minimize both the number of false positives and false negatives and maintain both a high precision and recall score. We attribute Deepdefacer's better recall and precision balance from training on a robust dataset that is rotated, scaled, and augmented that forces the model to predict more generalizable masks. Furthermore, having 92\% fewer parameters than the baseline model also reduces its tendency to overfit on intricate details within a scan, which makes it a more efficient model for binary segmentation. The quantitative results from both Figure \ref{fig:speed} and Table \ref{table:prdice} demonstrate that Deepdefacer is able to adequately balance masking quality with prediction speed on images with similar voxel intensities to those in the training set.

\subsection{Qualitative Image Comparison} 

\begin{figure}%
    \centering
    \subfloat[]{{\includegraphics[width=3.40cm]{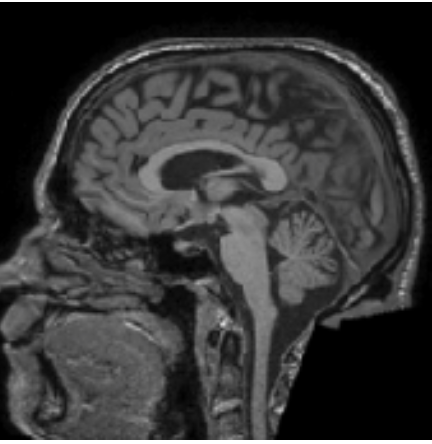} }}%
    \qquad
    \subfloat[]{{\includegraphics[width=3.40cm]{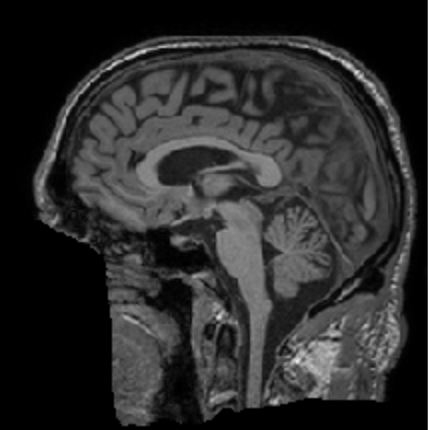} }}%

    \caption{(a) is a mislabeled output from the Pydeface program, while (b) is the same output from Deepdefacer. While Pydeface is unable to generalize well to all types of MRI images, Deepdefacer is still able to create accurate masks relying on facial patterns rather than intricate voxel details within an image. }%
    \label{fig:example}%
\end{figure}

\begin{figure}%
    \centering
    \subfloat[]{{\includegraphics[width=3.40cm]{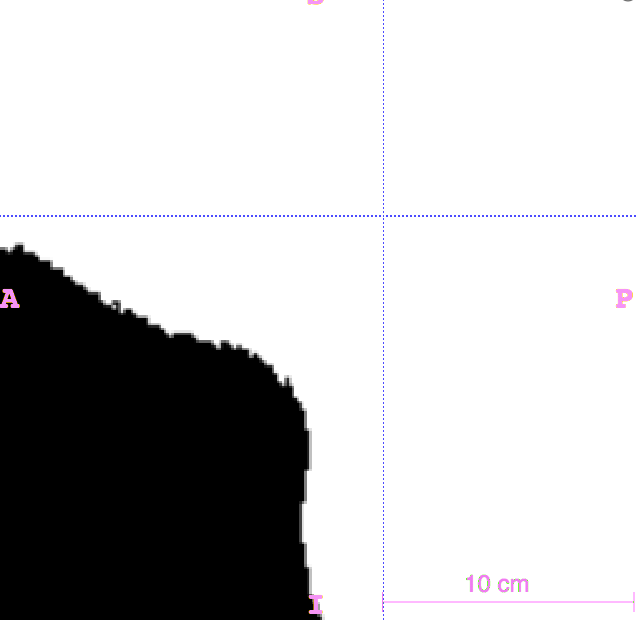} }}%
    \qquad
    \subfloat[]{{\includegraphics[width=3.40cm]{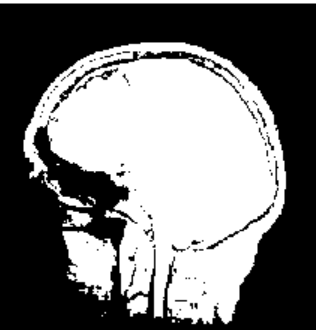} }}%

    \caption{(a) is a segmentation map produced by Deepdefacer (b) is a segmentation map produced by the 3D U-Net baseline. Axis in (a) follow the same definition listed under Figure 4. }%
    \label{fig:example}
\end{figure}

\begin{figure*}%
    \centering
    \subfloat[]{{\includegraphics[width=3.70cm]{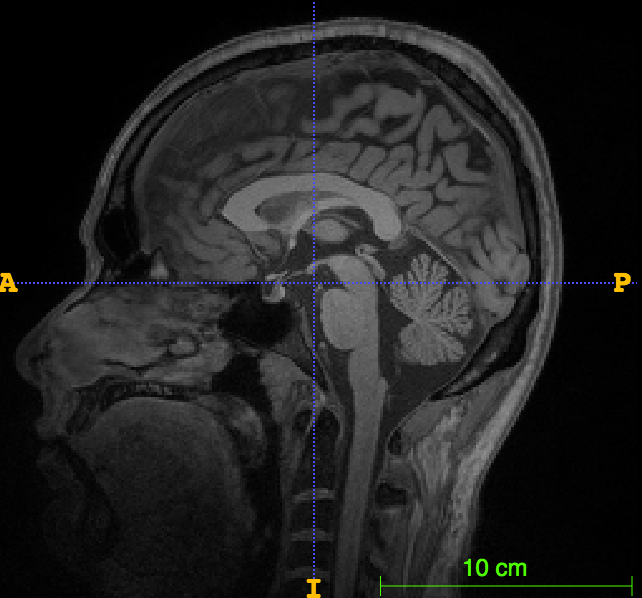} }}%
    \subfloat[]{{\includegraphics[width=3.70cm]{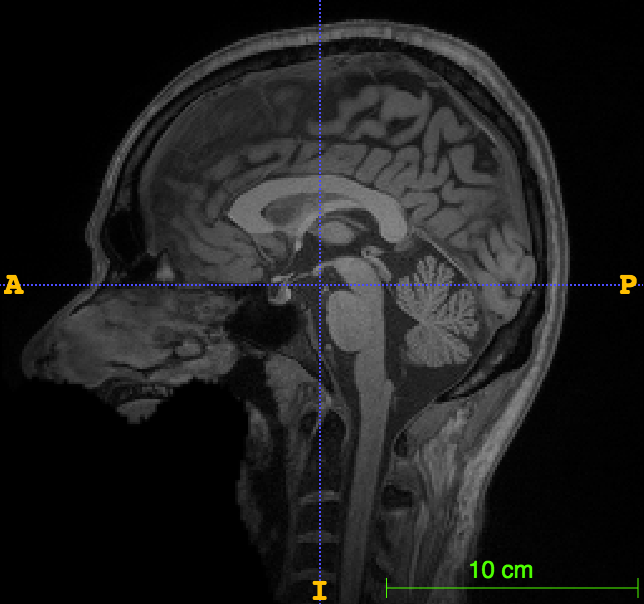} }}%
    \subfloat[]{{\includegraphics[width=3.70cm]{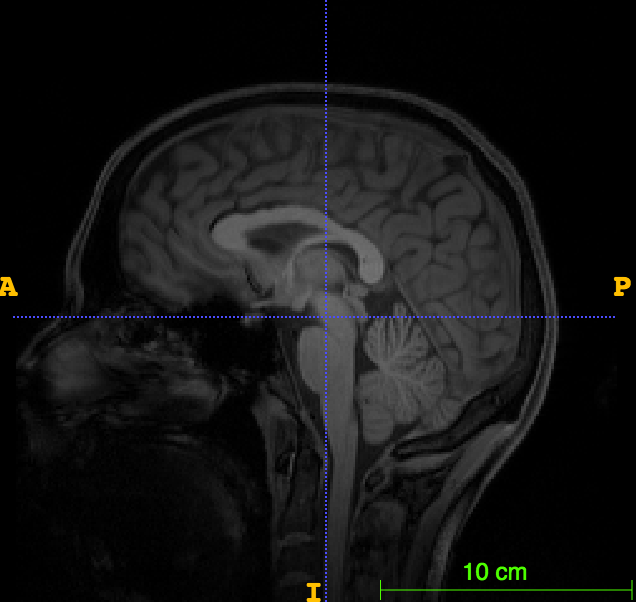} }}%
    \subfloat[]{{\includegraphics[width=3.70cm]{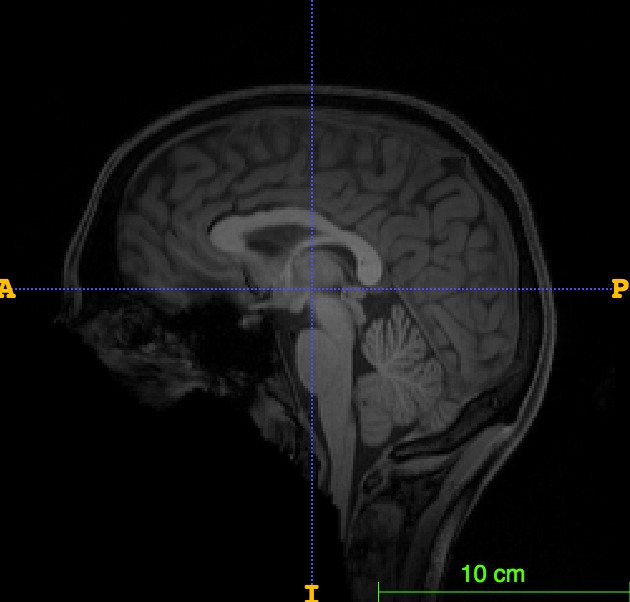} }}%
    \caption{Both (a) and (c) are images that do not appear in either the IXI or ICBM datasets, and have image resolution size and spacing compositions that are different from anything that Deepdefacer encounters in the training set. (b) and (d) are Deepdefacer's attempts at defacing (a) and (c) respectively. Axes in (a) - (d) follow the same definition listed under Figure 4. }%
    \label{fig:example}%
\end{figure*}
\vspace{-5pt}

While Pydeface supports between-modality co-registration to deal with different modalities, traditional defacing software can run into issues with images that have facial abnormalities. This disadvantage is apparent in Figure 7, where (a) is an example of an incorrectly defaced label produced by Pydeface, and (b) is Deepdefacer's output on the same exact image. Pydeface fails to mask this image due to computing an incorrect transformation matrix for translating its facemask onto the image, likely due to misinterpreting the overextended facial region in the scan. On the other hand, Deepdefacer produces a far superior defacing, which points to its ability to generalize to images with abnormalities that may not have been highly represented in the training set. This underscores how deep learning methods have a significant advantage over more traditional algorithms for anonymizing MRI images. 

One might point out that Deepdefacer's segmentation map in Figure 8a and defaced outputs in Figure 9 are amorphous rather than rigid like Pydeface's mask in Figure 5. This is likely due to Deepdefacer's smaller filter sizes throughout its network, which allows the model to better generalize to other types of MRI images. Furthermore, training on segmentation maps also forces Deepdefacer to avoid overfitting on features in a MRI image that are not in its facial region. Rigid or amorphous structure, however, is less important for determining the quality of an anonymized MRI image; the primary goal for this task is to just eliminate most of the facial region of a scan without perturbing voxels representing brain tissue. Thus, the masks's shape is irrelevant as long as the aforementioned objective is accomplished.

\section{Conclusion}

This paper presents Deepdefacer, an inaugural application of deep learning to MRI anonymization. We demonstrate its superiority over a 3D U-Net baseline model and the classical defacing software Pydeface, along with discussing the several benefits that deep learning has for solving problems in the anonymization space.

We see that Deepdefacer masks images up to 91\% faster on average than Pydeface and the baseline U-Net model on an array of CPU and GPU devices. The model is also able to deface images with a far smaller memory footprint of 92\% fewer parameters than the existing 3D U-Net architecture. Both of these advantages can potentially allow users to deface large MRI images on their own personal workstations, as opposed to relying on expensive cloud hardware or complicated defacing software. 

In conclusion, we hope this paper encourages other researchers to explore the many benefits of employing deep learning techniques in the MRI anonymization space. There are many possible extensions to this work including improving the performance of Deepdefacer on images of differing voxel intensities, delving into the interpretability of the model by analyzing correlation between the model weights and defacing predictions, and creating a web UI with a deepdefacer backend to let users deface images without requiring any deep learning or programming knowledge. We encourage interested readers to visit the github page at the end of this paper if they want to use or extend Deepdefacer's capabilities.

\section{Future Work}
As discussed in this paper, existing defacing
software is unnecessarily complex and requires installing several dependencies prior to usage.
Furthermore, this software is computationally expensive which bars researchers
who do not have access to expensive cloud infrastructure or personal machines.

We have released an experimental version of the model discussed in this paper via a PyPi program to hopefully address some of these obstacles, and also encourage other researchers interested in this work to iterate on its performance \cite{pypiprogram}. We also plan to integrate the model within a web application for an easy-to-use and efficient way to anonymize private MRI data. In order to accomplish this, we plan to package Deepdefacer within a tensorflow.js
module due to its ability to allow in-browser inferences, so we can ensure
that the researcher's data is private and not transmitted over the wire. Furthermore, we plan on allowing the user to opt-in and provide feedback on the accuracy of our model. With this feedback data, we can further improve the model's accuracy 
far beyond the capabilities of Pydeface (the ground truth model mentioned in this paper). Lastly, we hope to run Deepdefacer
over publicly hosted MRI data sets and provide our own anonymized MRI dataset that can be used by other researchers for their own tasks. 

\section{Contributions \& Acknowledgement}

We would like to thank the Poldrack Lab at the Stanford Center for Reproducible Neuroscience for providing our team with T1-Weighted MRI images for experimentation and advising us while working on this project. 

% Github Link for Deepdefacer:

% {\small \url{https://github.com/AKhazane/DeepDeface}}

\bibliography{main}
\bibliographystyle{plain}

\end{document}